\documentclass{article}

\usepackage{arxiv}

\usepackage[utf8]{inputenc} 
\usepackage[T1]{fontenc}    
\usepackage{hyperref}       
\usepackage{url}            
\usepackage{booktabs}       
\usepackage{amsfonts}       
\usepackage{nicefrac}       
\usepackage{microtype}      
\usepackage{lipsum}
\usepackage{graphicx}
\graphicspath{ {./images/} }

\usepackage[edges]{forest}
\usepackage{tikz}
\usepackage{xcolor} 

\title{From Snapshots to Symphonies: The Evolution of Protein Prediction from Static Structures to Generative Dynamics and Multimodal Interactions}

\author{
  Jingzhi Chen \\
  Shenzhen University of Advanced Technology \\
  \And
  Lijian Xu  \thanks{Corresponding author} \\
  Shenzhen University of Advanced Technology \\
  \texttt{xulijian@suat-sz.edu.cn}
}

\begin{document}
\maketitle
\begin{abstract}
The protein folding problem has been fundamentally transformed by artificial intelligence, evolving from static structure prediction toward the modeling of dynamic conformational ensembles and complex biomolecular interactions. This review systematically examines the paradigm shift in AI driven protein science across five interconnected dimensions: unified multimodal representations that integrate sequences, geometries, and textual knowledge; refinement of static prediction through MSA free architectures and all atom complex modeling; generative frameworks, including diffusion models and flow matching, that capture conformational distributions consistent with thermodynamic ensembles; prediction of heterogeneous interactions spanning protein ligand, protein nucleic acid, and protein protein complexes; and functional inference of fitness landscapes, mutational effects, and text guided property prediction. We critically analyze current bottlenecks, including data distribution biases, limited mechanistic interpretability, and the disconnect between geometric metrics and biophysical reality, while identifying future directions toward physically consistent generative models, multimodal foundation architectures, and experimental closed loop systems. This methodological transformation marks artificial intelligence's transition from a structural analysis tool into a universal simulator capable of understanding and ultimately rewriting the dynamic language of life.
\end{abstract}


\section{Introduction}
The milestone resolution of the protein folding problem marks the beginning of a new era in computational biology. AlphaFold 2 successfully elevated the prediction accuracy of monomeric protein structures to the atomic level~\cite{jumper_highly_2021}. This breakthrough not only addressed the fifty-year challenge of static folding but also significantly expanded the structural knowledge of protein sequence space through the construction of the AlphaFold Database~\cite{qu2024p}. However, static crystal structures represent only snapshots at specific energy minima and fail to fully explain the dynamic nature of proteins as they execute functions within complex cellular environments. 

As research advances, the focus of the field has inevitably shifted from single static structure prediction to more complex biomolecular interactions, conformational distributions, and unified multimodal modeling \cite{jumper_highly_2021, jing2024alphafold}.
This shift in research focus reflects a profound methodologicaltransformation in AI-driven scientific research. Regarding data representation, traditional prediction paradigms rely heavily on co-evolutionary information embedded in multiple sequence alignments~\cite{jumper_highly_2021}. While effective, this approach faces inherent limitations when handling orphan sequences or designed proteins and involves high computational costs. To address this bottleneck, the field has gradually moved toward utilizing protein language models to extract high-dimensional feature representations directly from single sequences \cite{zhou2025lm2protein, zheng2024esm}. The emergence of models such as ESM\_AA demonstrates that unified multi-scale modeling can capture physicochemical laws at both the residue and atomic levels without relying on homology alignment \cite{zheng2024esm}. This transition from explicit co-evolutionary information to implicit language model representations provides a foundation for addressing a broader range of biological questions.

In terms of the modeling mechanism, the evolution from discriminative models toward generative models constitutes another major trend. Traditional regression models tend to predict the mode of a conditional probability distribution, representing the most likely single structure, which often ignores the inherent thermodynamic fluctuations and conformational heterogeneity of proteins. The introduction of generative frameworks such as diffusion models and flow matching enables computational models to learn and sample protein conformational ensembles \cite{wang2024protein, jing2024alphafold}. For instance, works such as AlphaFlow and EigenFold embed prediction networks into generative frameworks, maintaining high-accuracy structure prediction while simulating conformational distributions consistent with molecular dynamics simulations \cite{jing2024alphafold}. This methodological innovation allows for the re-examination of the relationship between protein structure and function from a probabilistic perspective, bridging the gap between static structures and dynamic functions.

More importantly, proteins rarely act in isolation during life processes; their functional realization typically depends on specific interactions with other biomolecules. The release of AlphaFold 3 extends the prediction scope from pure protein chains to generalized biomolecular complexes including nucleic acids, small molecule ligands, and modified residues \cite{abramson_accurate_2024}. This enhancement in multimodal interaction prediction requires models to understand not only the folding rules of proteins themselves but also the physicochemical interaction laws between different biochemical entities. The parallel development of works such as RoseTTAFoldNA further confirms the feasibility and necessity of processing protein-nucleic acid complexes within a unified framework \cite{baek2024accurate}.
This paper aims to systematically review the progress of protein prediction technologies under this paradigm shift. To illustrate this rapid trajectory, Figure 1 maps the milestone architectures reviewed herein, highlighting the transition from foundational static structure prediction to advanced generative and multimodal frameworks.Unlike surveys focusing on inverse folding and de novo design, this work concentrates on the core task of prediction, covering the full spectrum from static structures to dynamic ensembles and from monomer folding to complex interactions. Although generative models like RFdiffusion are primarily applied in the design field \cite{butcher2025novo, watson_novo_2023}, their underlying geometric generative logic shares a deep dual relationship with prediction tasks. By analyzing these frontier methods, we attempt to answer how artificial intelligence evolves from a mere structural analysis tool into a general physical simulator for understanding the complex behaviors of biological macromolecules.

\begin{figure}[htbp]
    \centering
    \includegraphics[width=\textwidth]{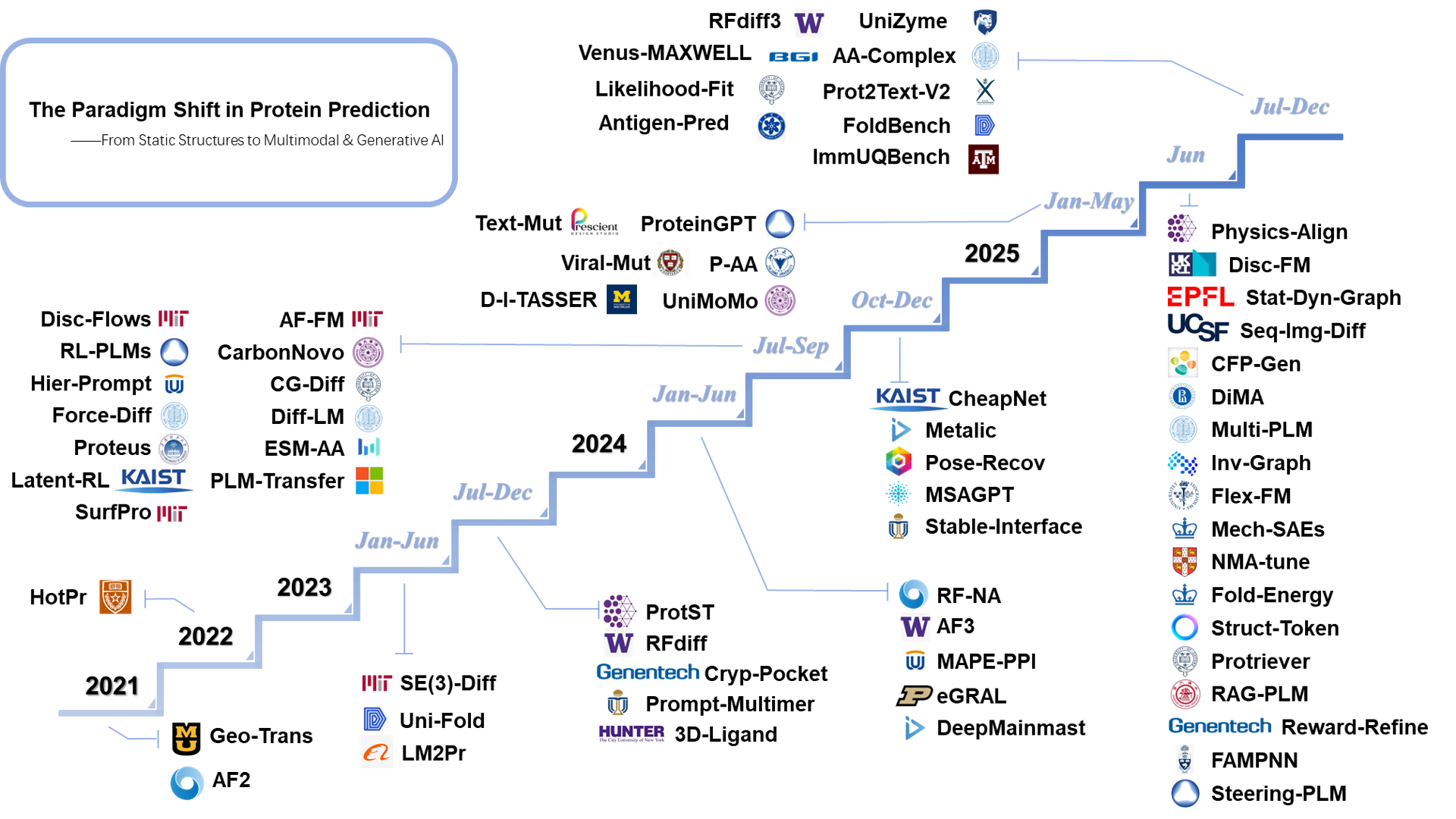}
    \caption{\textbf{Timeline of milestone AI models in protein prediction (2021–2025).} The chart illustrates the chronological progression of key literature reviewed in this survey. Models are positioned according to their publication or preprint release dates, capturing the rapid evolution from foundational structure prediction to recent advanced generative and multimodal frameworks. Logos indicate the primary affiliated institutions or companies.}
    \label{fig:timeline}
\end{figure}

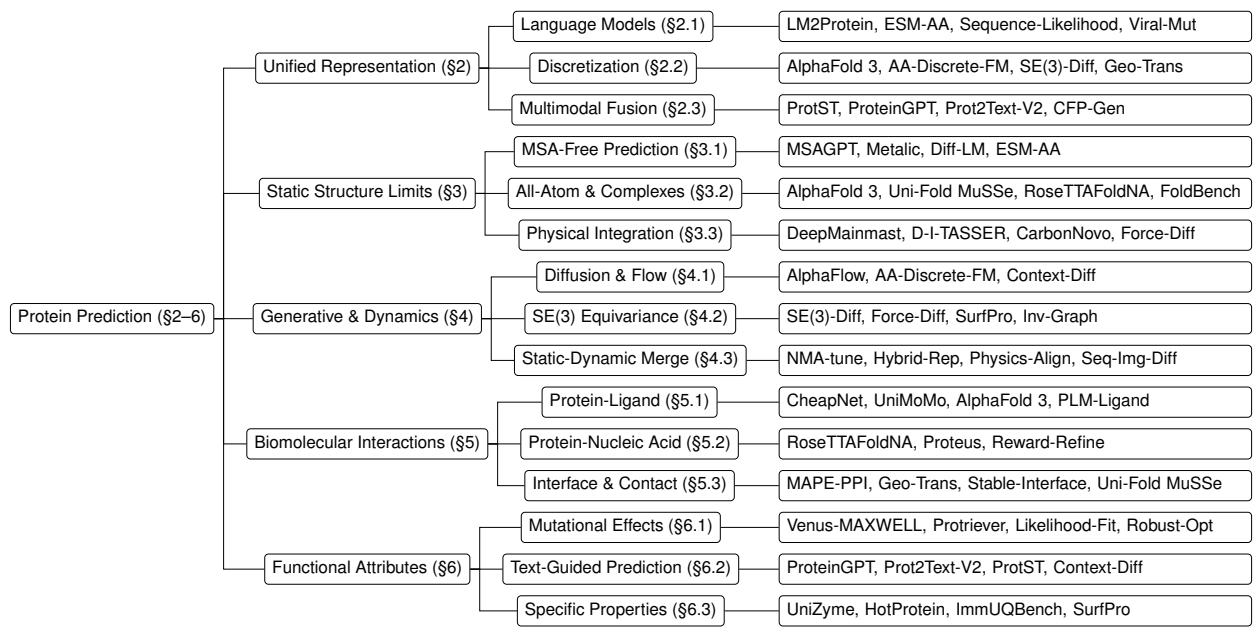
\begin{figure*}[htbp]
\centering
\resizebox{\textwidth}{!}{%
\begin{forest}
forked edges,
for tree={
  grow'=0,
  draw,
  rounded corners=2pt,
  font=\sffamily\footnotesize,
  inner ysep=3pt,
  inner xsep=4pt,
  s sep=2mm,
  l sep=6mm,
},
where n children=0{
  tier=leaf,
  text width=8.5cm,   
  align=center,
}{
  align=center,
}
[
  {Protein Prediction (\S 2--6)}
  [
    {Unified Representation (\S 2)}
    [
      {Language Models (\S 2.1)}
      [{LM2Protein, ESM-AA, Sequence-Likelihood, Viral-Mut}]
    ]
    [
      {Discretization (\S 2.2)}
      [{AlphaFold 3, AA-Discrete-FM, SE(3)-Diff, Geo-Trans}]
    ]
    [
      {Multimodal Fusion (\S 2.3)}
      [{ProtST, ProteinGPT, Prot2Text-V2, CFP-Gen}]
    ]
  ]
  [
    {Static Structure Limits (\S 3)}
    [
      {MSA-Free Prediction (\S 3.1)}
      [{MSAGPT, Metalic, Diff-LM, ESM-AA}]
    ]
    [
      {All-Atom \& Complexes (\S 3.2)}
      [{AlphaFold 3, Uni-Fold MuSSe, RoseTTAFoldNA, FoldBench}]
    ]
    [
      {Physical Integration (\S 3.3)}
      [{DeepMainmast, D-I-TASSER, CarbonNovo, Force-Diff}]
    ]
  ]
  [
    {Generative \& Dynamics (\S 4)}
    [
      {Diffusion \& Flow (\S 4.1)}
      [{AlphaFlow, AA-Discrete-FM, Context-Diff}]
    ]
    [
      {SE(3) Equivariance (\S 4.2)}
      [{SE(3)-Diff, Force-Diff, SurfPro, Inv-Graph}]
    ]
    [
      {Static-Dynamic Merge (\S 4.3)}
      [{NMA-tune, Hybrid-Rep, Physics-Align, Seq-Img-Diff}]
    ]
  ]
  [
    {Biomolecular Interactions (\S 5)}
    [
      {Protein-Ligand (\S 5.1)}
      [{CheapNet, UniMoMo, AlphaFold 3, PLM-Ligand}]
    ]
    [
      {Protein-Nucleic Acid (\S 5.2)}
      [{RoseTTAFoldNA, Proteus, Reward-Refine}]
    ]
    [
      {Interface \& Contact (\S 5.3)}
      [{MAPE-PPI, Geo-Trans, Stable-Interface, Uni-Fold MuSSe}]
    ]
  ]
  [
    {Functional Attributes (\S 6)}
    [
      {Mutational Effects (\S 6.1)}
      [{Venus-MAXWELL, Protriever, Likelihood-Fit, Robust-Opt}]
    ]
    [
      {Text-Guided Prediction (\S 6.2)}
      [{ProteinGPT, Prot2Text-V2, ProtST, Context-Diff}]
    ]
    [
      {Specific Properties (\S 6.3)}
      [{UniZyme, HotProtein, ImmUQBench, SurfPro}]
    ]
  ]
]
\end{forest}%
}
\caption{\textbf{A comprehensive taxonomy of AI architectures in protein prediction.} 
The landscape is organized into five primary dimensions: representation learning (\S 2), 
static structure refinement (\S 3), generative dynamics (\S 4), biomolecular interactions (\S 5), 
and functional attribute prediction (\S 6). 
\textbf{Note:} Model names are shown without citations for clarity and rendering stability; 
full references are provided in the main text.}
\label{fig:taxonomy}
\end{figure*}
\section{Unified Representation: From Sequence to All-Atom}
A core challenge in protein science is mapping complex biophysical entities into mathematical objects suitable for computation. This process involves more than encoding one-dimensional amino acid sequences; it must encompass three-dimensional geometric structures, dynamic conformational distributions, and high-dimensional semantic functional descriptions. Effective representation learning serves as the bridge between biological laws and deep learning algorithms, determining whether a model can capture profound physicochemical essences under limited supervision. This chapter explores the technical evolution from sequence language models toward geometry-aware and multimodal fusion representations.  
\subsection{Protein Language Models}
Amino acid sequences carry historical records of billions of years of evolution, with their arrangement patterns implying folding rules and functional constraints. Traditional computational methods rely heavily on multiple sequence alignments to extract co-evolutionary signals to infer contact relationships between residues. However, AlphaFold 2 demonstrated that while co-evolutionary information is vital, it has limitations when processing orphan sequences or design tasks \cite{jumper_highly_2021}.To overcome this bottleneck, the field has turned to the protein language model paradigm using self-supervised learning on massive unlabeled sequences.

The core hypothesis of this paradigm is that by training on large-scale corpora to predict masked residues or the next token, models can implicitly learn the contextual dependencies and physicochemical properties of amino acids. Research by Li et al. reveals the phenomenon of feature reuse in such large-scale pre-training, proving that representations learned by language models can be directly transferred to downstream tasks such as structure prediction and functional annotation \cite{li2024feature}. As parameter scales increase, models exhibit emergent abilities; for example, LM2Protein and ESM All-Atom demonstrate through unified multi-scale modeling that single-sequence models possess the potential to rival alignment-based methods in capturing atomic-level details \cite{zhou2025lm2protein, zheng2024esm}.
Beyond structural information, pre-trained representations contain rich fitness landscape information. Gurev et al. successfully predicted the mutational effects of viral proteins using sequence models without relying on expensive experimental data \cite{gurev2025sequence}. Further research indicates a strong correlation between model likelihood and protein fitness, with methods proposed by Pugh et al. directly utilizing these probability distributions to infer the survival advantages of variants \cite{pugh2025likelihood}. In more complex scenarios, works such as Venus-MAXWELL and Protriever demonstrate how combining retrieval mechanisms with language models can enhance the prediction of mutational stability by searching for homologous sequences in latent space \cite{weitzman2025protriever, yu2025venus}. Additionally, for predicting specific biochemical functions, breakthroughs like UniZyme in cleavage site recognition further confirm that language models can capture subtle specific recognition patterns between enzymes and substrates \cite{li2025unizyme}. These advances collectively indicate that protein language models have become a universal mathematical foundation for understanding sequence-function relationships.

\subsection{Discretization of Structure and Microenvironment}
Although sequences contain all folding information, function is often directly determined by three-dimensional structures and local microenvironments. Three-dimensional coordinates are continuous geometric data with rotational and translational invariance, which creates a natural modal mismatch with the Transformer architecture proficient at processing discrete symbols. To resolve this, the tokenization of structural data has become a key technology connecting geometric and semantic spaces.
A mainstream strategy involves quantifying local backbone structures into a discrete vocabulary. Benchmarking by Yuan et al. provides a detailed evaluation of different structural tokenization schemes, noting that reasonable discretization significantly reduces the complexity of geometric generation \cite{yuan2025protein}. 

Building on this, the generative flow models proposed by Campbell et al. demonstrate how to jointly design sequences and structures through multimodal flow matching on discrete state spaces \cite{campbell2024generative}. Yi et al. further developed an all-atom discrete flow matching algorithm, proving the effectiveness of discretized representations in capturing side-chain conformational details through inverse folding tasks \cite{yi2025all}.
To ensure representations adhere to the geometric laws of physical space, $SE(3)$ equivariance must be introduced. Force-guided diffusion models proposed by Wang et al. and the $SE(3)$ diffusion framework by Yim et al. strictly guarantee the geometric consistency of predicted structures under coordinate transformations by defining the generative process on Lie groups \cite{wang2024protein, yim2023se}. Such geometric priors are crucial for capturing interactions at protein interfaces. Geometric Transformers by Morehead et al. and atomic-level graph neural networks by Fiorellini-Bernardis et al. significantly improve contact map prediction accuracy by explicitly encoding the relative positions and orientations between nodes \cite{morehead2021geometric}.
Furthermore, protein function is often determined by local microenvironments such as active sites. The microenvironment-aware prompt learning strategy proposed by Wu et al. enhances the ability of models to predict protein-protein interactions by focusing on the spatial features of residue neighborhoods \cite{wu2024learning, wu2024mape}. Work by Widatalla et al. also emphasizes the importance of side-chain conditional modeling in all-atom design, indicating that refined microenvironment representation is a necessary condition for achieving high-precision structure generation \cite{widatalla2025sidechain}. The success of AlphaFold 3 pushes this all-atom contextual representation to its limit, achieving joint modeling of unified systems involving proteins, nucleic acids, and ligands \cite{abramson_accurate_2024}.
\subsection{Multimodal Fusion}

The complete picture of a protein includes not only physical entities but also accumulated textual knowledge, image data, and interaction relationships within biological networks. Representations of a single modality struggle to cover this high-level semantic information; thus, constructing aligned multimodal spaces has become a frontier direction.    
To align natural language with biological sequences, the ProtST framework proposed by Xu et al. uses contrastive learning to bridge the distance between protein sequences and biomedical text in the feature space \cite{xu2023protst}. This idea is further extended in ProteinGPT and Prot2Text-V2, multimodal large models that not only understand sequence patterns but also generate functional descriptions or predict structural properties based on textual instructions \cite{xiao2024proteingpt, fei2025prot2text}. Berenberg et al. explored using text conditioning to guide mutational effect prediction, enabling models to reason based on specific phenotypes described in literature \cite{berenberg2025residue}.

Knowledge graphs, as carriers of structured knowledge, provide explicit logical constraints for models. Retrieval-augmented language models proposed by Zhang et al. enhance the richness and accuracy of representations by integrating entity relationships from external knowledge bases \cite{zhang2024retrieval}. Similarly, Wang et al. use knowledge-aware reinforcement learning strategies to guide directed evolution paths for proteins, ensuring that generated variants are biologically plausible \cite{wang2024knowledge}. Regarding the image modality, Zheng et al. attempted to link protein sequences with microscopy images using diffusion models, offering new possibilities for inferring molecular structures from cellular phenotypes \cite{zheng2025bridging}.

The ultimate goal of multimodal fusion is to achieve cross-modal generation and design. Hsieh et al. systematically elaborated on the design space of multimodal protein language models, pointing out that fusing information from different modalities can significantly improve the success rate of generative tasks \cite{hsieh2025elucidating}. Works such as UniMoMo and CFP-Gen demonstrate how to design novel proteins with specific properties under a unified generative framework using joint constraints from text, structure, and functional labels \cite{kong2025unimomo, yin2025cfp} . This unified representation, blending sequence, geometry, and semantics, marks that artificial intelligence is moving from simple data fitting toward a comprehensive understanding and creation of biological mechanisms.

Similar cross-modal alignment strategies have proven successful in vision-language models, including occlusion-based contrastive learning with semantic-aware views \cite{yang2025one,feng2026efficient} and unified instruction tuning for medical image interpretation \cite{xu2023learning}. 
Recent progress in efficient visual representation learning further reinforces the feasibility of compact yet expressive multimodal representations. Methods such as adaptive token basis learning that reduces redundant visual tokens \cite{young2026fewer} and robust spatial-concept alignment strategies \cite{young2026scalar} demonstrate that high-level semantic features can be preserved even under substantial feature compression. These principles have also shown effectiveness in domain-specific scenarios including medical image understanding \cite{yang2024segmentation,yang2023geometry}, pathology token compression \cite{chen2026tc,wu2026towards}, multimodal medical foundation models \cite{xu2024medvilam,xu2024foundation}, and vision-centric long-context compression frameworks \cite{gao2026zerosense,he2026autoselect}. Collectively, these advances indicate that carefully designed token representations can maintain discriminative power while significantly reducing computational redundancy, further supporting the development of unified capacity-aware representations across multimodal learning systems.

\section{Refining the Limits of Static Structure Prediction}
Following the establishment of high-accuracy benchmarks for monomeric protein structure prediction by AlphaFold 2, the research focus of the field has shifted toward addressing residual bottlenecks. Current optimization paths concentrate on three primary dimensions: eliminating the costly dependency on multiple sequence alignments (MSA) to accommodate high-throughput and orphan protein prediction, transcending the single-chain limitation to achieve multi-molecular complex assembly with all-atom precision, and integrating physical energy and experimental constraints to correct biophysical biases inherent in purely data-driven models.
\subsection{Single-Sequence and MSA-Free Prediction}
While co-evolutionary information derived from multiple sequence alignments is vital for inferring resid-residue contacts, the construction process is computationally expensive and often ineffective for orphan proteins or synthetic designs that lack homologous sequences. To overcome this limitation, researchers have explored utilizing protein language models to mine implicit structural features within single sequences. Methods proposed by Fang et al. demonstrate that mapping three-dimensional coordinates directly from pre-trained language model embeddings achieves considerable prediction accuracy without explicit homology retrieval [50]. 

Subsequent works such as LM2Protein and ESM All-Atom further demonstrate that single-sequence models possess the potential to rival alignment-based methods in capturing atomic-level details through increased parameter scales and multi-scale training \cite{zhou2025lm2protein, zheng2024esm}.
To compensate for the absence of evolutionary information in alignment-free scenarios, MSAGPT introduces a neural prompting mechanism that uses generative pre-training to simulate the distributional characteristics of MSAs, thereby generating virtual co-evolutionary contexts during inference without requiring retrieval \cite{chen2024msagpt}. For low-resource scenarios, the Metalic framework adopts a meta-learning strategy that enables models to rapidly adapt to new protein families through few-shot in-context learning, significantly reducing the reliance on large-scale evolutionary data \cite{beck2024metalic}. The effectiveness of these MSA-free methods is theoretically supported by research from Li et al. regarding feature reuse in large-scale pre-training, while the application of sparse autoencoders elucidates how language models implicitly encode structural semantics such as helices and sheets \cite{li2024feature, adams2025mechanistic, parsan2025towards}. Furthermore, diffusion language models serve as versatile sequence learners and exhibit robust generalization capabilities across diverse protein distributions \cite{wang2024diffusion, meshchaninov2024diffusion}.

\subsection{All-Atom Precision and Complex Prediction}
Biological functions frequently depend on the synergistic effects of multi-protein complexes and intricate interactions with other biomolecules. The introduction of AlphaFold 3 marks an expansion of prediction scope from monomeric chains to biomolecular systems with all-atom precision, utilizing a unified diffusion architecture to process the joint folding of proteins, nucleic acids, small molecule ligands, and modified residues \cite{abramson_accurate_2024}. Following this trend, models such as Uni-Fold MuSSe address the challenges of modeling inter-chain interactions in multimer prediction by introducing multimodal sequence encoding and cross-chain attention mechanisms \cite{zhu2023uni}. Gao et al. explore prompt-based strategies that guide pre-trained models to capture complex subunit assembly rules through specific task prompts \cite{gao2024protein}. 
The accurate prediction of protein interaction interfaces is essential for complex assembly. Geometric Transformers proposed by Morehead et al. significantly improve contact map prediction accuracy by explicitly encoding the relative positions and orientations of interface regions \cite{morehead2021geometric}. To address instability in interface prediction, Gao et al. introduce contrastive learning to enhance the robustness of interface representations, while MAPE-PPI utilizes microenvironment-aware embeddings for the efficient construction of large-scale interaction networks \cite{wu2024mape, gao_towards_2024}. Within the context of all-atom generation, RFdiffusion3 and related all-atom generative models demonstrate the feasibility of designing and predicting biomolecular interactions at atomic resolution, a generative capability that in turn validates the profound understanding of side-chain physicochemical properties by the prediction models \cite{butcher2025novo, chen2025all, qu2024p, widatalla2025sidechain}.    

The boundaries of prediction have also extended to non-protein molecules. RoseTTAFoldNA addresses the gap in protein-nucleic acid complex prediction \cite{baek2024accurate}, while CheapNet and PLM-based ligand binding site prediction models demonstrate efficient prediction of binding modes and affinities between proteins and small molecules in drug discovery scenarios \cite{lim2025cheapnet, lampsite2023}. The establishment of benchmarks such as FoldBench provides a standardized metric system for evaluating the comprehensive performance of these all-atom, multimodal models \cite{xu2025benchmarking}.
\subsection{Integrating Physical and Experimental Data}
Purely data-driven deep learning models occasionally produce structural hallucinations that violate physical principles, such as atomic overlaps or anomalous bond lengths. To correct these biases, the integration of physical energy and experimental constraints serves as a crucial corrective measure. Protocols such as DeepMainmast demonstrate how cryo-electron microscopy (cryo-EM) density maps can be integrated as external constraints into deep learning workflows, enabling the reconstruction of high-confidence atomic models even from low-resolution data \cite{terashi2024deepmainmast}. Zheng et al. further explore the use of diffusion models to link protein sequences with microscopy images, utilizing macroscopic phenotypic data to constrain microscopic structure prediction \cite{zheng2025bridging}.

Regarding physical energy guidance, D-I-TASSER combines deep learning potential energy surfaces with iterative physical simulations to ensure the thermodynamic stability of predicted structures \cite{zheng2025deep}. The force-guided diffusion model proposed by Wang et al. incorporates the gradient of physical force fields into the generative process, ensuring that each sampling step proceeds along the direction of energy descent \cite{wang2024protein}. Similarly, Lu et al. emphasize the role of physical feedback in aligning conformational generation with true distributions \cite{lu2025aligning}. This hybrid modeling strategy is equally applicable to functional prediction, where works like Venus-MAXWELL and CarbonNovo achieve precise inference of mutational effects and structural stability by learning protein fitness landscapes and energy distributions \cite{ren2024carbonnovo, deng2025predicting, yu2025venus}. Robust optimization methods proposed by Lee et al. further prove that combining physical constraints in latent space effectively avoids local extrema in fitness landscapes \cite{lee2024robust}. These advances indicate that future prediction systems will become organic unifications of data pattern recognition and physical law constraints.  

\begin{figure}[htbp]
    \centering
    \includegraphics[width=1.0\textwidth]{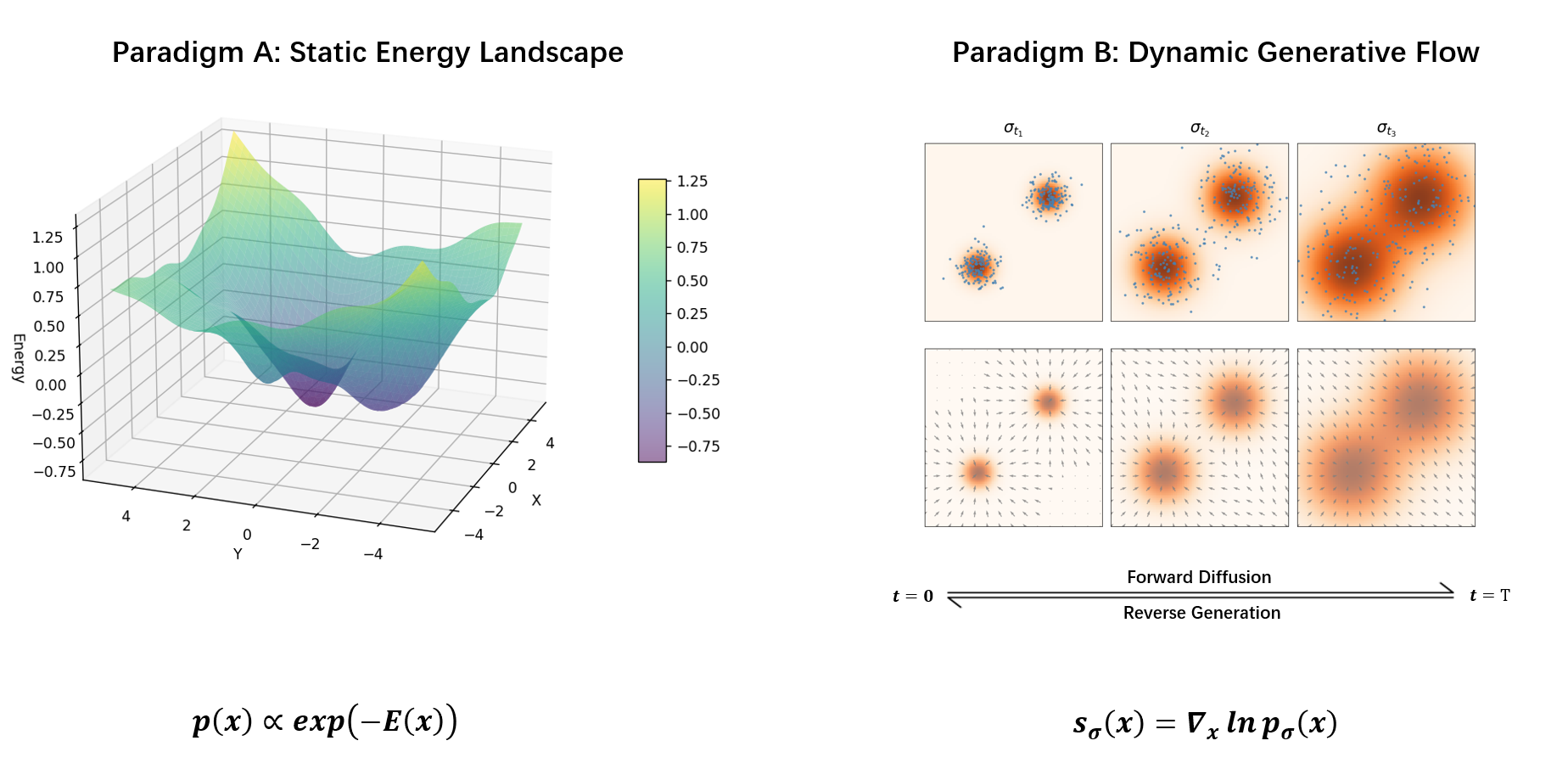} 
    
    \caption{Evolution of Generative Paradigms: From Static Energy Landscapes to Multi-scale Dynamic Score Flows. 
    (a) Paradigm A illustrates Energy-Based Models (EBMs) characterized by a global scalar field $E(\mathbf{x})$. 
    (b) Paradigm B depicts the Score-based framework. The forward process perturbs data via a noise schedule $\sigma(t)$, while the reverse flow follows the learned score field $\mathbf{s}_{\sigma}(\mathbf{x})$ to recover the data manifold. This shift from global density to local score-based guidance ($\sigma \propto t$) resolves the intractability of the partition function $Z$.}
    \label{fig:paradigm_comparison}
    \label{fig:paradigm_shift}
\end{figure}
\section{Generative Prediction: Capturing Conformational Ensembles and Dynamics}
Proteins are not static rigid bodies; their functions often rely on dynamic transitions across complex free-energy surfaces. Traditional prediction paradigms tend to output a single static structure, which obscures the rich dynamical properties of proteins in physiological environments. Recent advances in artificial intelligence, particularly the introduction of generative models, allow us to transcend the limitations of a single optimal solution and instead simulate the Boltzmann distribution of proteins to capture their conformational ensembles and dynamical behavior.

\subsection{Diffusion Models and Flow Matching}
Diffusion models generate data by reversing a noise process, a characteristic that makes them inherently suitable for simulating protein folding pathways from disorder to order. Unlike traditional regression models aimed at finding global energy minima, generative models focus on learning the probability distribution of conformations. Breakthrough works such as AlphaFlow and EigenFold apply flow matching algorithms to protein structure prediction, successfully generating conformational ensembles highly consistent with molecular dynamics simulations \cite{jing2024alphafold}. These models no longer output a unique crystal structure but instead sample a series of structural variants that adhere to thermodynamic statistical laws, thereby revealing the conformational heterogeneity of proteins across different functional states. Koo et al. further demonstrate that by learning from molecular dynamics trajectories, deep learning models can directly predict conformational ensembles that reflect real physical fluctuations \cite{koo2025learning}.

Beyond continuous geometric spaces, generative models in discrete spaces also exhibit significant potential. Wang et al. and Meshchaninov et al. respectively explore the versatility of diffusion language models in sequence space, proving that diffusion in latent space effectively captures the distribution patterns of amino acid sequences \cite{wang2024diffusion, meshchaninov2024diffusion}. The all-atom discrete flow matching algorithm proposed by Yi et al. validates the effectiveness of discretized representations in capturing subtle side-chain conformations through inverse folding tasks \cite{yi2025all}. Additionally, the context-guided diffusion strategy proposed by Klarner et al. significantly enhances model generalization on out-of-distribution data, enabling the handling of unseen protein families \cite{klarner2024context}. Collectively, these works indicate that generative models are not only tools for design but also prediction engines for understanding complex protein distribution characteristics.

\subsection{Geometric Flows and SE(3) Equivariance}
To ensure that generated structures adhere to the geometric laws of physical space, models must satisfy SE(3) equivariance, where predicted results transform accordingly with rotations and translations of the coordinate system. The SE(3) diffusion model proposed by Yim et al. establishes a solid geometric foundation for backbone generation, strictly guaranteeing the self-consistency of generated trajectories in physical space \cite{yim2023se}. Building upon this, Wang et al. introduce a force-guided diffusion mechanism that integrates the gradients of physical force fields into the generative process, ensuring that sampling steps follow the direction of energy descent and achieving an organic integration of physical-driven and data-driven approaches \cite{wang2024protein}.
Such geometric priors are especially important for processing complex biomolecular systems. AlphaFold 3 achieves joint geometric modeling of proteins, nucleic acids, and ligand complexes by constructing a unified diffusion architecture, demonstrating the ability to capture intermolecular interactions at an all-atom scale \cite{abramson_accurate_2024}. For functional design involving continuous surfaces, the SurfPro model proposed by Song et al. utilizes sophisticated geometric representations to capture refined features of protein surfaces \cite{song2024surfpro}. Furthermore, Wang et al. enhance graph contrastive learning from the perspective of invariance, further improving model robustness against geometric structural perturbations \cite{wang2025enhancing}.
\subsection{Merging Static and Dynamic Perspectives}
A comprehensive description of protein physical properties requires merging static topological structures with dynamic fluctuation information. Guo et al. propose a hybrid representation strategy that simultaneously encodes static protein contact maps and dynamic motion modes, significantly enhancing the model ability to capture functional movements \cite{guo2025boosting}. NMA-tune by Komorowska et al. utilizes low-frequency vibration information generated by normal mode analysis to fine-tune generative models, ensuring that generated backbone structures are not only geometrically sound but also possess the designability required for specific functions \cite{komorowska2025nma}.
Physical feedback plays a critical role in connecting static generation with dynamic distributions. Lu et al. discuss how physical feedback can be used to align generated conformational ensembles with true physical distributions \cite{lu2025aligning}. This introduction of a dynamic perspective is essential for understanding cryptic binding sites; research by Bloore et al. suggests that only by fully understanding the pocket exposure mechanisms during dynamic conformational changes can one accurately predict pharmacophore binding sites that appear only in specific states \cite{bloore2023protein}. Zheng et al. attempt to link protein sequences with microscopy images using diffusion models, employing macroscopic phenotypic data to constrain microscopic structure prediction and providing new insights for cross-scale dynamic modeling \cite{zheng2025bridging}. These fusion strategies ensure that prediction models are no longer limited to static snapshots but can present a dynamic panorama of proteins in life processes.
\section{Beyond Proteins: Predicting Biomolecular Interactions}
Life processes are complex molecular symphonies where proteins rarely perform in isolation. Moving from monomeric prediction to the prediction of complexes and heterogeneous molecular interactions is an inevitable stage in the development of computational biology. This chapter discusses how artificial intelligence handles specific recognition and complex assembly between proteins and ligands, nucleic acids, and other proteins.
\subsection{Protein-Ligand and Small Molecule Interactions}
Accurately predicting the binding sites and affinities between proteins and small molecules is a central objective of drug discovery. The release of AlphaFold 3 extends the scope of prediction to biomolecular systems with all-atom precision. By employing a unified diffusion architecture, this model enables the direct prediction of protein-ligand complex structures and demonstrates a strong capacity to generalize across the geometric features of non-protein molecules \cite{abramson_accurate_2024}. Regarding binding site prediction, PLM-driven methods proposed by Zhang et al. show that deep semantic features captured by protein language models can precisely localize potential small molecule binding regions \cite{lampsite2023}. Furthermore, research by Bloore et al. reveals that by accounting for the dynamic conformational changes of proteins, language models can identify cryptic pockets that are only exposed in specific transient structures, an ability that is essential for designing drugs against difficult-to-drug targets \cite{bloore2023protein}. Beyond structure prediction, the assessment of binding affinity is equally critical. CheapNet utilizes a hierarchical cross-attention mechanism to significantly reduce the computational cost of predicting protein-ligand binding affinity while maintaining high accuracy \cite{lim2025cheapnet}. Work by Errington et al. emphasizes the importance of evaluating the physical plausibility of predicted binding poses, noting that high-precision geometric reconstruction does not always correspond to an energetic optimum \cite{errington2025assessing}. From the dual perspective of generative design, UniMoMo demonstrates the utility of unified generative models in designing novel 3D molecules that bind specifically to target proteins. This advancement in inverse generation capability further validates the deep understanding of intermolecular interaction laws within prediction models \cite{kong2025unimomo}.
\subsection{Protein-Nucleic Acid Interactions}
The regulation of gene expression depends on the precise recognition between proteins and DNA or RNA. The introduction of RoseTTAFoldNA addresses the previous gap in nucleic acid complex prediction by providing an end-to-end approach to predict the three-dimensional structures of protein-nucleic acid complexes, which serves as a powerful tool for understanding the binding mechanisms of transcription factors \cite{baek2024accurate}. This breakthrough is also integrated into the unified framework of AlphaFold 3, further enhancing the accuracy of multimodal complex prediction \cite{abramson_accurate_2024}. In the field of design, models such as Proteus explore the universality of protein structure generation, while Uehara et al. demonstrate the introduction of reward-guided iterative optimization strategies within diffusion models to design proteins with specific DNA-binding specificities \cite{wang2024proteus, uehara2025reward}. These developments help analyze existing biological regulatory networks and provide theoretical guidance for the development of gene-editing tools in synthetic biology.
\subsection{Interface and Contact Prediction}
The accurate identification of protein-protein interaction interfaces is a prerequisite for complex assembly. The Geometric Transformer proposed by Morehead et al. significantly improves contact map prediction accuracy by explicitly encoding the relative positions and orientations of interface regions \cite{morehead2021geometric}. To address instability in interface prediction, Gao et al. propose a stable interface representation learning framework that enhances model sensitivity to the microenvironment of interface residues through contrastive learning \cite{gao_towards_2024}. Additionally, MAPE-PPI introduces microenvironment-aware embedding technology that integrates sequence context and local physicochemical features to achieve the rapid construction of large-scale PPI networks \cite{wu2024mape}.
In the context of complex structure prediction, models such as Uni-Fold MuSSe effectively solve the problem of modeling inter-chain interactions in multimer prediction by introducing multimodal sequence encoding and cross-chain attention mechanisms \cite{zhu2023uni}. The all-atom generative model proposed by Chen et al. further demonstrates the feasibility of designing and predicting protein complexes at atomic resolution \cite{chen2025all}. This progression from interface identification to all-atom assembly indicates that artificial intelligence is gradually mastering the complex sociological rules governing biological macromolecules.

\section{Functional Attributes and Fitness Prediction}
The ultimate significance of biological structure lies in its function. As representation capabilities improve, artificial intelligence is evolving from a simple structure prediction tool into a reasoning engine for functional attributes. This chapter focuses on fitness landscape prediction based on sequence and structure, text-guided functional annotation, and the evaluation of specific biophysical properties, showcasing the zero-shot generalization potential of deep learning models.

\subsection{Mutational Effects and Fitness Landscapes}
Predicting the impact of amino acid mutations on protein function and stability, a concept known as the fitness landscape, is a core task in protein engineering. Two primary prediction paradigms currently exist: sequence-based likelihood inference and structure-based energy calculation. The method proposed by Pugh et al. belongs to the former, calculating the generation probability of mutant sequences via pre-trained language models to directly translate likelihood scores into fitness indicators. This approach is computationally efficient and particularly suitable for high-throughput screening \cite{pugh2025likelihood}. In contrast, Deng et al. and Sharma et al. adopt structure-based methods that infer mutational effects by calculating changes in folding energy or structural stability scores. Although these methods involve higher computational costs, they provide clearer physical interpretability \cite{deng2025predicting, sharma2025exploring}.

To combine the advantages of both approaches, Venus-MAXWELL utilizes large-scale protein language models to efficiently learn protein-mutation stability landscapes, achieving precise prediction of mutational effects under zero-shot conditions \cite{yu2025venus}. For specific scenarios such as viral evolution, Gurev et al. demonstrate that sequence models alone can accurately track the fitness trajectories of viral proteins \cite{gurev2025sequence}. In more complex search tasks, Protriever enables end-to-end differentiable homology search and fitness prediction by retrieving homologous sequences in latent space, which effectively addresses generalization challenges caused by data scarcity \cite{weitzman2025protriever}. Furthermore, text conditioning is being introduced into mutation prediction. Berenberg et al. prove that using natural language descriptions as conditions can guide models to more accurately predict mutational effects under specific phenotypes \cite{berenberg2025residue}. For the optimization of fitness landscapes, the robust optimization strategy proposed by Lee et al. leverages the geometric properties of latent space to effectively avoid local extrema, providing a new computational path for directed evolution \cite{lee2024robust}.
\subsection{Text-Guided Functional Prediction and Question Answering}
The introduction of natural language processing technology allows for the direct extraction or generation of protein functional information from textual descriptions, marking a leap for AI from numerical prediction to semantic understanding. As a representative multimodal large model, ProteinGPT achieves the ability to generate functional descriptions or even predict three-dimensional structures based on text prompts by aligning protein sequence space with biomedical text space \cite{xiao2024proteingpt}. Prot2Text-V2, proposed by Fei et al., further enhances this cross-modal alignment. Through contrastive learning, the model understands abstract biological concepts such as kinase activity or membrane localization and maps them precisely onto sequence features \cite{fei2025prot2text}.
Knowledge enhancement is the key to improving text-based prediction accuracy. The ProtST framework developed by Xu et al. significantly improves model performance in downstream functional annotation tasks by integrating large-scale biomedical literature \cite{xu2023protst}. The retrieval-augmented language model proposed by Zhang et al. utilizes entity relationships in external knowledge bases to help the model maintain biological factual consistency when generating functional descriptions \cite{zhang2024retrieval}. In generative design, Klarner et al. use context-guided diffusion models combined with semantic information to guide the generation of novel proteins, ensuring that designed molecules are not only structurally sound but also consistent with intended functional descriptions \cite{klarner2024context}.
\subsection{Specific Property Prediction}
In addition to general functional prediction, attribute prediction models for specific industrial or medical needs are developing rapidly. In the field of enzyme engineering, UniZyme establishes a unified prediction framework across enzyme classes by fusing prior knowledge of enzyme active sites. It accurately identifies cleavage sites of unseen enzymes, demonstrating the significant potential of combining domain knowledge with deep learning \cite{li2025unizyme}. For thermal stability, a critical engineering attribute, the HotProtein framework focuses on learning temperature-related structural features, providing an effective guidance tool for designing heat-resistant proteins \cite{chen2022hotprotein}.
In immunological applications, Zai et al. improve the accuracy of vaccine antigen prediction by combining geometric deep learning with protein language models \cite{zai2025integrating}. Considering the reliability of prediction results, ImmUQBench establishes a benchmark for uncertainty quantification to systematically evaluate model confidence in immunogenicity prediction, which is essential for reducing clinical trial risks \cite{qayyum2026immuqbench}. Furthermore, works such as SurfPro capture the geometric and chemical features of protein surfaces to achieve precise identification of functional pockets, providing direct support for structure-based drug design \cite{song2024surfpro}. These specialized models for specific attributes complement general foundation models and together constitute the toolbox of modern computational biology.

\begin{figure*}[htbp]
  \centering
  \includegraphics[width=\textwidth]{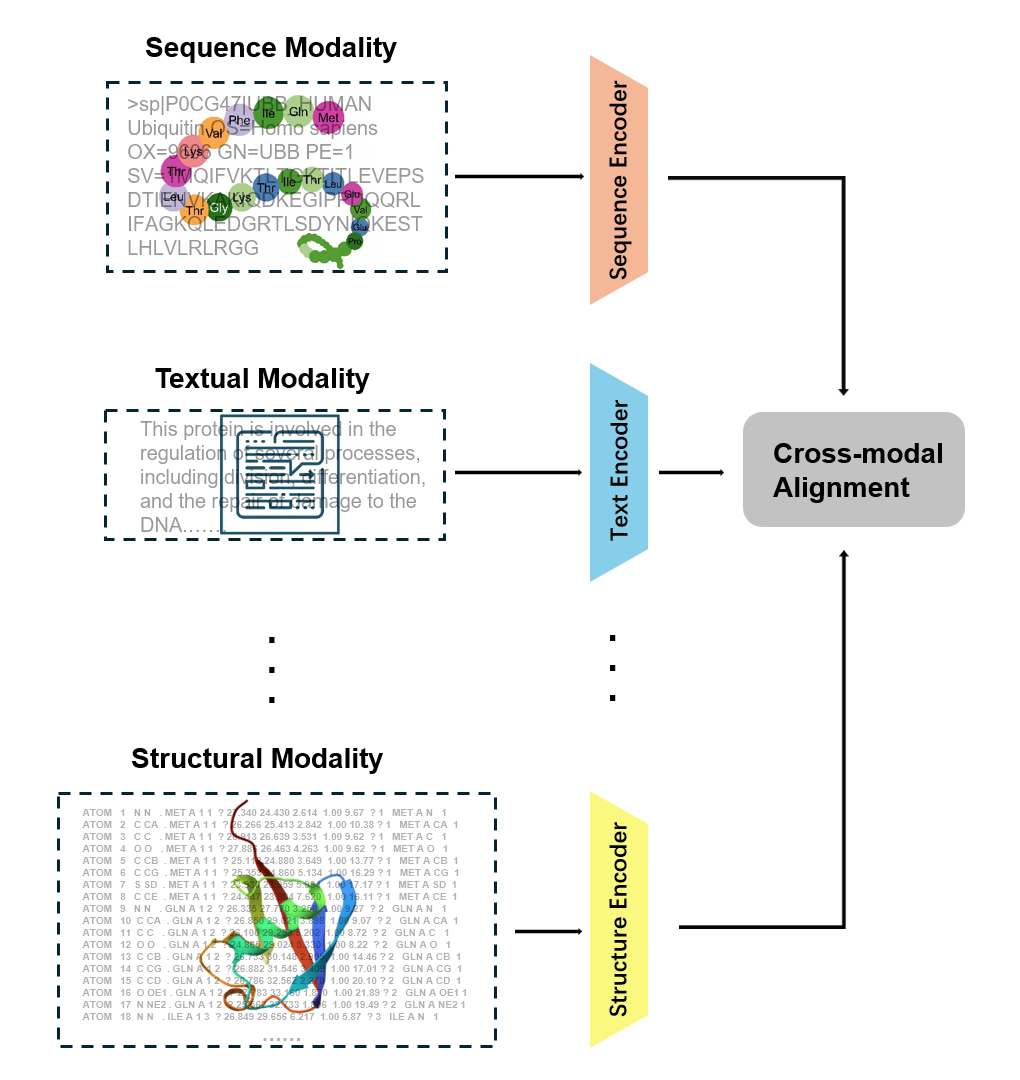} 
  \caption{\textbf{A typical multimodal integration architecture for protein representation learning.} By mapping diverse biological signals into an aligned latent space, this framework bridges the gap between low-level geometric constraints and high-level semantic knowledge, facilitating the transition from complex assembly to functional attribute prediction.}
  \label{fig:multimodal_alignment}
\end{figure*}
\section{Challenges and Future Work}
Despite the remarkable achievements of artificial intelligence in the field of protein prediction, even demonstrating the potential to exceed experimental precision in specific tasks, the transition from algorithmic prototypes to broad scientific applications still faces several fundamental bottlenecks. Examining these limitations not only facilitates an objective evaluation of current progress but also points the way for the development of next-generation biological foundation models. This chapter systematically analyzes the limitations of existing technologies and explores potential pathways toward understanding the language of life.
\subsection{Current Bottlenecks and Limitations}
\subsubsection{Data Boundaries and Distribution Biases}
Existing deep learning models depend heavily on established experimental databases, which limits their predictive capacity largely to the distribution of known protein families. A substantial gap exists between experimentally acquired structural data and the actual biological space. This non-uniformity of data distribution leads to a significant performance drop when models encounter orphan proteins or samples lacking homologous sequences \cite{weitzman2025protriever, beck2024metalic}. In rapidly evolving scenarios such as viral evolution, the problem of data scarcity is particularly acute, directly constraining the accuracy with which models capture mutational effects \cite{gurev2025sequence}.
Furthermore, current training data primarily originate from purified crystal structures in vitro, leading to an over-representation of standard conformations in these models. Benchmarks such as FoldBench reveal that models exhibit significant biases when processing non-standard residues and proteins in complex environments, reflecting a lack of perception of real biological environmental factors \cite{xu2025benchmarking}. As Varadi et al. note, the surge in AI-predicted data raises the risk that if future models are recursively trained on large-scale unverified predictions, systemic biases may accumulate \cite{qu2024p}. Functional features that depend on transient conformations, such as cryptic binding pockets, are extremely under-represented in existing datasets, making it difficult for models to accurately reconstruct the true dynamics of proteins within complex cellular environments \cite{bloore2023protein, qayyum2026immuqbench}.

\subsubsection{Lack of Mechanistic Interpretability}
The black-box nature of deep learning models remains a major obstacle to scientific trust. Although researchers observe improvements in prediction accuracy, it is often difficult to deconstruct the internal reasoning logic. This epistemological gap makes it impossible to determine whether a model has truly mastered physicochemical laws or is simply memorizing statistical features through complex pattern recognition. Research by Adams et al. and Parsan et al. attempts to deconstruct protein language models using sparse autoencoders, aiming to map high-dimensional hidden representations onto understandable biological semantics such as specific secondary structure motifs or functional site features \cite{adams2025mechanistic, parsan2025towards}.
Feature reuse analysis by Li et al. reveals the mechanism by which models extract existing knowledge during transfer learning, providing a preliminary perspective on the efficacy of pre-trained models \cite{li2024feature, huang2025steering}. However, most current interpretability studies are limited to a posteriori analysis and lack causal verification of the decision-making process. Although some works attempt to enhance transparency through generative simulations or the introduction of physical priors, these efforts have not yet fully eliminated the skepticism arising from the black-box effect \cite{chen2024msagpt, li2025unizyme}.
\subsubsection{Disconnection Between Evaluation Metrics and Biological Reality}
For a long time, the evaluation of prediction quality has relied heavily on geometric similarity metrics. This single-dimensional assessment system is increasingly showing its limitations. Errington et al. point out that in molecular docking and complex prediction, the precise reconstruction of geometric poses is not equivalent to the correct recovery of physical interactions \cite{errington2025assessing}. A predicted structure with low geometric deviation may exist in an unstable state on the energy surface or contain atomic clashes at critical interaction interfaces \cite{xu2025benchmarking}. This disconnection between geometric accuracy and biophysical reality reflects a desensitization of current loss functions toward chemical details.
With the rise of generative models, evaluating the accuracy of dynamic conformational ensembles presents a new challenge. Traditional snapshot comparison metrics cannot effectively measure the overlap between a generated distribution and a true thermodynamic ensemble, making it difficult to quantify progress in capturing conformational heterogeneity \cite{jing2024alphafold}. Additionally, the lack of a robust uncertainty quantification system limits the practical application of models in high-risk areas such as clinical trials or drug development \cite{qayyum2026immuqbench, yuan2025protein}.
\subsection{Future Research Directions}
\subsubsection{Toward Fully Dynamic Ensembles and Physical Consistency}
Since traditional geometric metrics cannot satisfy the requirements of dynamic assessment, the focus of future research must shift from the regression of single structures to the sampling alignment of ensemble distributions. By introducing flow matching algorithms and physical feedback mechanisms, models will be able to simulate the dynamic transitions of proteins across complex free-energy surfaces, thereby accurately reconstructing the probability distribution of conformations \cite{jing2024alphafold, viliuga2025flexibility}. Utilizing physical priors such as normal mode analysis to guide the generative process ensures that predicted structures are not only spatially reasonable but also possess the designability required for functional movements \cite{komorowska2025nma, lu2025aligning}. Refining this dynamic perspective will significantly enhance the ability to capture functional motion features such as cryptic pockets \cite{bloore2023protein}.
\subsubsection{Constructing Multimodal General Foundation Models}
Future work will focus on building general artificial intelligence capable of understanding the language of life across all scales. By integrating sequences, structures, text, and microscopy images into a unified generative framework, models will possess cross-modal reasoning capabilities \cite{hsieh2025elucidating, zheng2024esm}. Deep semantic alignment technologies will enable models to perform functional understanding and attribute prediction based on natural language instructions, much like human experts \cite{xiao2024proteingpt, fei2025prot2text, xu2023protst, berenberg2025residue}. The integrated application of cross-scale data, such as linking protein sequences with microscopy images at subcellular resolution, will provide new possibilities for understanding how molecular behavior translates into macroscopic phenotypes \cite{zheng2025bridging, kong2025unimomo}.

\subsubsection{Experimental Closed-Loop and Active Learning Optimization}
The ultimate value of artificial intelligence in biological discovery lies in establishing an automated scientific closed-loop. By tightly coupling prediction models with high-throughput wet-lab experiments and using knowledge-aware reinforcement learning strategies to guide directed evolution, the iterative optimization of protein functions can be achieved rapidly \cite{wang2024knowledge}. This active learning paradigm allows models to continuously explore the latent space and self-correct through experimental feedback, effectively avoiding hallucinatory regions in the fitness landscape \cite{lee2024robust, campbell2024generative, yin2025cfp}. Combined with fast task-transfer techniques under few-shot conditions, future prediction systems will be able to adapt quickly to new biological challenges with minimal data costs \cite{beck2024metalic, klarner2024context}.

\section{Conclusions}

This paper systematically reviews the evolution of protein prediction—from static structure resolution to dynamic conformational generation and multimodal interaction prediction. A profound methodological transformation is evident: the field has shifted from isolated geometric modeling toward a comprehensive understanding of biophysical principles.
The role of AI in this process is undergoing a qualitative change. Early discriminative models served as assistive tools for structural biology. With the advent of protein language models and generative architectures, computational approaches now demonstrate the capacity to simulate complex physical distributions. AI is gradually evolving into a universal simulator for the language of life, bridging the gap between sequence and function while capturing atomic-scale protein dynamics.

However, despite rapid progress, the field confronts fundamental challenges including data bias, limited interpretability, and imperfect evaluation metrics. Constructing physically consistent generative models that integrate physical priors with deep learning represents a critical pathway forward. Establishing automated closed-loop systems between prediction and experimentation will further accelerate our understanding of life's mechanisms.
Looking ahead, protein prediction will converge with protein design. As multimodal foundation models mature, we anticipate unified architectures capable of integrating sequences, structures, text, and omics data. This multi scale understanding will enable AI not only to read structural snapshots of life but also to write dynamic biological functions, bringing disruptive breakthroughs to drug discovery and synthetic biology. In summary, AI driven protein science is transitioning from reproducing known structures toward understanding the fundamental laws of life, opening a new era for reshaping the biological world.

\clearpage

\bibliographystyle{unsrt}  
\bibliography{references}

\end{document}